# GAMBIT+: A Challenge Set for Evaluating Gender Bias in Machine Translation Quality Estimation Metrics

Giorgos Filandrianos[1,2,*]    Orfeas Menis Mastromichalakis[1]    Wafaa Mohammed[3]
Giuseppe Attanasio[2]    Chrysoula Zerva[2,4,5]
[1]National Technical University of Athens, Greece
[2]Instituto de Telecomunicações, Lisbon, Portugal    [3]University of Amsterdam, Netherlands
[4]Instituto Superior Técnico, Universidade de Lisboa, Portugal    [5]ELLIS Unit Lisbon, Portugal

## Abstract

Gender bias in machine translation (MT) systems has been extensively documented, but bias in automatic quality estimation (QE) metrics remains comparatively underexplored. Existing studies suggest that QE metrics can also exhibit gender bias, yet most analyses are limited by small datasets, narrow occupational coverage, and restricted language variety. To address this gap, we introduce a large-scale challenge set specifically designed to probe the behavior of QE metrics when evaluating translations containing gender-ambiguous occupational terms. Building on the GAMBIT corpus of English texts with gender-ambiguous occupations, we extend coverage to three source languages that are genderless or natural-gendered, and eleven target languages with grammatical gender, resulting in 33 source–target language pairs. Each source text is paired with two target versions differing only in the grammatical gender of the occupational term(s) (masculine vs. feminine), with all dependent grammatical elements adjusted accordingly. An unbiased QE metric should assign equal or near-equal scores to both versions. The dataset's scale, breadth, and fully parallel design, where the same set of texts is aligned across all languages, enables fine-grained bias analysis by occupation and systematic comparisons across languages.

## 1 Introduction

While gender bias in machine translation systems is widely acknowledged and has been widely documented, the biases in the translations of gender-ambiguous terms, such as occupational titles, is a topic relatively unexplored, which has lately gained some traction (Mastromichalakis et al., 2025). These biases are often reflected in the disproportionate assignment of a gender (e.g. masculine forms) to occupations when the gender of the subject is unknown. While overall much research has focused on bias in MT outputs, comparatively little attention has been paid to the potential gender biases of automatic quality estimation metrics. Recent studies suggest that QE metrics, which are intended to provide an objective measure of translation quality, can also exhibit systematic biases (Zaranis et al., 2024). However, existing analyses on gender-ambiguous inputs are typically limited by datasets with short texts, typically sentence-level, a restricted variety of occupations, and a limited range of language pairs.

To address these limitations, we introduce GAMBIT+[1], a large-scale challenge set specifically designed to evaluate gender bias in QE metrics when translating gender-ambiguous occupational terms. Our dataset extends the original GAMBIT corpus[2] (Mastromichalakis et al., 2025) of English texts with gender-ambiguous occupations to include three source languages—English (a natural-gendered language), Turkish and Finnish (both genderless languages)—and eleven target languages with grammatical gender: Arabic, Czech, Greek, Spanish, French, Icelandic, Italian, Portuguese, Russian, Serbian, and Ukrainian. This results in 33 source–target language pairs, covering a wide spectrum of linguistic typologies and providing a rich resource for cross-linguistic analysis of gender bias in QE.

For each source text, we provide two target versions differing only in the grammatical gender of the occupational term (masculine vs. feminine). All necessary adjustments for grammatical correctness, such as gendered adjectives, are applied consistently across both versions. An unbiased QE metric should assign equal or near-equal scores to the two versions, as there is no indication of gender in the source text. Unlike prior work, which often aggregates bias analysis across all occupations or texts in general, GAMBIT+ explicitly tracks the occupational terms mentioned in each text and links them to ISCO-08 codes (hereafter, ISCO codes) [3]. This enables fine-grained, occupation-level

---

[*]Corresponding author: geofila@ails.ece.ntua.gr.

[1]GAMBIT+ is available at https://huggingface.co/datasets/ailsntua/gambit-plus.
[2]https://huggingface.co/datasets/ailsntua/GAMBIT
[3]ISCO-08 is an internationally recognized system for

analyses and facilitates the study of stereotypical patterns, as previous studies have shown that certain occupations are more likely to be translated in a gendered manner according to societal stereotypes (Mastromichalakis et al., 2025; Menis-Mastromichalakis et al., 2025).

The parallel design of the dataset ensures that all source texts are aligned across target languages and both gendered versions, enabling consistent and controlled benchmarking of QE metrics. With this resource, researchers can investigate not only overall tendencies of QE systems but also nuanced, occupation-specific, and cross-lingual patterns, providing a more comprehensive understanding of how gender bias manifests in translation evaluation.

As part of the challenge set subtask of the shared task on Automated Translation Quality Evaluation Systems at WMT 2025 (Lavie et al., 2025), GAMBIT+ was used to evaluate a set of established QE metrics as well as participant submissions to the shared task. This allowed us to benchmark the performance of different metrics in a controlled, fine-grained setting and to assess how gender bias manifests in real-world QE systems. In this paper, we present and discuss the results of these evaluations, highlighting both systematic tendencies and occupation-specific patterns, and providing insights into the current limitations and strengths of existing QE approaches with respect to gender fairness.

## 2 Related Work

Gender bias in Machine Translation has been widely documented, revealing persistent disparities influenced by societal norms, model design choices, and deployment contexts (Savoldi et al., 2021; Vanmassenhove, 2024; Savoldi et al., 2024; Menis-Mastromichalakis et al., 2025; Mastromichalakis et al., 2025). Numerous studies have examined the prevalence and consequences of such biases across languages, cultural settings, and MT architectures (Rescigno et al., 2020; Paolucci et al., 2023; Farkas and Németh, 2022; Ghosh and Caliskan, 2023; Kostikova et al., 2023; Piazzolla et al., 2023), underscoring the need for robust evaluation and mitigation. A particularly relevant line of work focuses on occupational bias in MT, where stereotypes associated with specific professions affect translation choices

---

classifying occupations endorsed by the International Labour Organisation (ILO). It provides a hierarchical structure that categorizes jobs into four levels of increasing granularity, using a digit-based coding system. More details: https://ilostat.ilo.org/methods/concepts-and-definitions/classification-occupation/

(Gorti et al., 2024; Tal et al., 2022; Mastromichalakis et al., 2025). Related efforts in other NLP tasks have explored gender ambiguity in Question Answering (Parrish et al., 2022; Li et al., 2020) and coreference resolution (Rudinger et al., 2018; Zhao et al., 2018; Kotek et al., 2023). In MT, proposed strategies for handling ambiguity include generating all grammatically correct gendered translations (Garg et al., 2024), and disambiguating inputs before translation (Vanmassenhove et al., 2018). Resources supporting these investigations range from knowledge graphs (Mastromichalakis et al., 2024) to multilingual benchmarks and challenge sets (Currey et al., 2022). Mitigation strategies have included model fine-tuning, data balancing, and adaptive learning (Saunders and Byrne, 2020; Escudé Font and Costa-jussà, 2019; Costa-jussà and de Jorge, 2020), as well as gender-neutral translation approaches (Piergentili et al., 2023a; Lardelli and Gromann, 2023) and benchmarks for evaluating them (Piergentili et al., 2023b; Lardelli et al., 2024; Gkovedarou et al., 2025). A central element in all these efforts is evaluation, since quality estimation metrics determine what counts as a "good" translation and thus influence MT system development.

While several studies have examined whether evaluation metrics for natural language generation exhibit social biases—such as Qiu et al. (2023), who compared n-gram- and model-based metrics, Sun et al. (2022), who quantified different bias types, and Gao and Wan (2022), who measured race and gender stereotypes—very few works have explored gender bias in MT QE metrics. One exception is Zaranis et al. (2024), which conducted a multifaceted analysis of QE methods and demonstrated that the metrics themselves can be biased, potentially perpetuating the very stereotypes they are used to assess.

A key limitation of existing work on gender bias in QE is the lack of suitable datasets, particularly for studying gender ambiguity. Many challenge sets, such as WinoMT (Stanovsky et al., 2019) and MT-GenEval (Currey et al., 2022), focus on cases where gender can be resolved via coreference or other contextual cues. The MuST-SHE corpus (Bentivogli et al., 2020) similarly contains audio and textual cues that reveal gender (e.g., speaker's voice, pronouns, named entities). Conversely, datasets containing true gender ambiguity (Rudinger et al., 2018; Zhao et al., 2018) are often limited in language coverage, consist of isolated sentences, and lack occupational diversity, restricting fine-grained analysis. In contrast, our challenge set builds on GAMBIT (Mastromichalakis et al., 2025), where occupational mentions are inherently gender-ambiguous and no gold-standard gendered translation exists. This enables us to test whether QE metrics assign different scores to two theoretically equivalent translations differing only in the grammatical gender of the occupation, offering a controlled setting for investigating metric bias in ambiguous contexts.

## 3 Challenge Set Creation

The creation of the challenge set, GAMBIT+, was grounded in the GAMBIT dataset (Mastromichalakis et al., 2025), which covers all ISCO occupations (all 4-digit codes) and contains the corresponding ISCO codes and names, with English texts distributed evenly across five formats: short stories, brief news reports, short statements, short conversations, and short presentations, in which the given occupation appears. In these texts, the gender of the occupation is not explicitly indicated. For example, in the sentence "The professor delivered an engaging lecture on generation modification by the auditorium", shown in Table 1, the occupation "professor" is mentioned without any linguistic cues that would reveal its gender.

In GAMBIT+, these gender-neutral English sentences were translated into gendered target languages, producing parallel masculine and feminine versions of each sentence while preserving all other aspects of the translation. In addition, we extended the source language coverage to include genderless languages such as Turkish and Finnish, alongside English, ensuring a diverse and balanced dataset. In total, the challenge set comprises 29,415 source instances (9,805 each of the three source languages), which were translated into 11 target languages, resulting in 323,565 triplets, each consisting of the source sentence with a masculine and a feminine translation. Table 1 presents an example from the GAMBIT+ dataset with the data from GAMBIT and the corresponding entries created for the Challenge Set. Specifically, the following sections describe the generation process for producing these translations and the evaluation procedure applied to assess and ensure the high quality of the resulting dataset.

### 3.1 Generation

**Gendered Languages.** The construction of the challenge set for gendered languages was based on the English texts from the GAMBIT dataset in which the gender of the occupation was ambiguous. These sentences served as the source material for the initial stage of the process. An LLM[4] was subsequently instructed to translate each sentence into the target language, specifically translating the occupation into a given gender form (e.g., masculine). Following this, a separate interaction with the same model was initiated, ensuring that no conversational history from the first stage was preserved, and the model was instructed to take the translated text and produce a variation in the target language that preserved the text exactly, modifying only the gender of the occupation from the initial to the alternate form (e.g., feminine respectively).

This process yielded three aligned versions for each source sentence:

1. the original English text $s$ containing a gender-ambiguous occupation,

2. the translation with the occupation in its masculine form $t_{male}$, and

3. the translation with the occupation in its feminine form $t_{female}$.

The masculine and feminine translations were constructed to be semantically and lexically identical, differing only in the gender marking of the occupation and the dependent grammatical elements. There were a few cases where the masculine and feminine variations were identical in some target languages, mostly due to the morphology of the source text and the grammatical restrictions of the target languages. All such examples were discarded from all languages (so if the masculine and feminine translations were identical in at least one language, this text was removed from all languages) in order to keep the challenge set consistent, and the texts in all languages aligned. After filtering, we retained 8,771 samples per language pair, resulting in a total of 289,443 samples in the GAMBIT+ dataset.

**Genderless Languages.** For genderless languages, a similar approach was applied. In this case, the model was instructed to translate the original text into the genderless target language without introducing any linguistic cues that might imply a specific gender for the occupation. This step was designed to avoid subtle cases where unintended gender hints could arise, even in languages without grammatical gender. Since in those languages there is no grammatical gender, the second step that generated the gendered variation in gendered languages was not applied.

By following this procedure, we obtained aligned texts across multiple languages, comprising source languages with gender-ambiguous occupations and target languages with matched masculine and feminine translations for the same occupations, as shown

---
[4]The LLM used is Claude Sonnet 3.5 v2

| GAMBIT | | | |
| --- | --- | --- | --- |
| **ISCO ID** | **ISCO Name** | **Type** | **English Text** |
| 2310 | Professor | short statement | The Professor delivered an engaging lecture on quantum mechanics to a packed auditorium. |

| GAMBIT+ Additional Source Languages | |
| --- | --- |
| **Language** | **Text** |
| Turkish | Profesör, tıklım tıklım dolu bir konferans salonunda kuantum mekaniği üzerine etkileyici bir ders verdi. |
| Finnish | Professori piti mukaansatempaavan luennon kvanttimekaniikasta täydelle luentosalille. |

| GAMBIT+ Target Languages | | |
| --- | --- | --- |
| **Language** | **Masculine** | **Feminine** |
| Arabic | ألقى الأستاذ محاضرة مشوقة عن الميكانيكا الكمية في قاعة محاضرات مكتظة. | ألقت الأستاذة محاضرة مشوقة عن الميكانيكا الكمية في قاعة محاضرات مكتظة. |
| Greek | Ο Καθηγητής έδωσε μια συναρπαστική διάλεξη για την κβαντική μηχανική σε ένα κατάμεστο αμφιθέατρο. | Η Καθηγήτρια έδωσε μια συναρπαστική διάλεξη για την κβαντική μηχανική σε ένα κατάμεστο αμφιθέατρο. |
| Czech | Profesor přednesl poutavou přednášku o kvantové mechanice před zaplněným auditoriem. | Profesorka přednesla poutavou přednášku o kvantové mechanice před zaplněným auditoriem. |
| Icelandic | Herra prófessorinn flutti áhugaverðan fyrirlestur um skammtafræði fyrir fullum fyrirlestrasal. | Frú prófessorinn flutti áhugaverðan fyrirlestur um skammtafræði fyrir fullum fyrirlestrasal. |
| Italian | Il Professore ha tenuto una coinvolgente lezione sulla meccanica quantistica in un auditorium gremito. | La Professoressa ha tenuto una coinvolgente lezione sulla meccanica quantistica in un auditorium gremito. |
| Russian | Профессор прочитал увлекательную лекцию по квантовой механике в переполненной аудитории. | Профессор прочитала увлекательную лекцию по квантовой механике в переполненной аудитории. |
| French | Le Professeur a donné une conférence passionnante sur la mécanique quantique dans un amphithéâtre comble. | La Professeure a donné une conférence passionnante sur la mécanique quantique dans un amphithéâtre comble. |
| Spanish | El Profesor dio una conferencia cautivadora sobre mecánica cuántica ante un auditorio repleto. | La Profesora dio una conferencia cautivadora sobre mecánica cuántica ante un auditorio repleto. |
| Portuguese | O Professor ministrou uma palestra envolvente sobre mecânica quântica para um auditório lotado. | A Professora ministrou uma palestra envolvente sobre mecânica quântica para um auditório lotado. |
| Serbian | Profesor je održao zanimljivo predavanje o kvantnoj mehanici pred punim auditorijumom. | Profesorica je održala zanimljivo predavanje o kvantnoj mehanici pred punim auditorijumom. |
| Ukrainian | Професор провів захопливу лекцію з квантової механіки в переповненій аудиторії. | Професорка провела захопливу лекцію з квантової механіки в переповненій аудиторії. |

Table 1: Example instance from GAMBIT+ dataset, showing GAMBIT metadata, source translations, and target translations (masculine/feminine).

in the example of Table 1. The model used for the generation is Claude Sonnet 3.5 v2 [5].

## 3.2 Evaluation

Ensuring that the translations differed solely in the gender marking of the occupation, while remaining otherwise semantically equivalent, was a central requirement of the data creation process. However, adherence to this constraint could not be assumed, even though the model had been explicitly prompted to preserve all other aspects of the source text. To verify compliance, an additional evaluation was conducted in an LLM-as-a-judge approach (Gu et al., 2024). A different and more capable model[6], namely Claude Sonnet 3.7 [7], was provided with two texts: the translation in the masculine form and the translation in the feminine form. The model was instructed to analyse the two sentences and determine whether the only difference between them was the gender of the occupation, or whether additional semantic or structural differences were present. This evaluation was carried out for all language-pair combinations using 10% of the dataset due to computation constraints, and the results are reported in Table 2.

As we can see, in almost all cases the accuracy is above 90%, indicating a high quality of generated data. Additionally, we manually reviewed around 100 samples of the "errors" identified by the LLM judge to further investigate the issues. In most cases, the errors flagged by the judge were not truly errors but rather stemmed from minor variations in dependent gram-matical elements, which should indeed be adjusted to match the gender changes of the occupation. Such differences often led the judge to conclude that the outputs were not aligned, revealing a certain oversensitivity in the evaluation. As a result, when the metric marked something as incorrect, it was not necessarily a substantive error, whereas we observed that when it marked something as correct, this judgment was indeed accurate. Consequently, the reported numbers are likely stricter than the actual performance.

---
[5] anthropic.claude-3-5-sonnet-20241022-v2:0
[6] https://www.anthropic.com/news/claude-3-7-sonnet
[7] anthropic.claude-3-7-sonnet-20250219-v1:0

| Lang. | Acc. % | Lang. | Acc. % |
|---|---|---|---|
| Portuguese | 98.38 | Russian | 94.00 |
| Spanish | 97.62 | Ukrainian | 92.62 |
| French | 97.25 | Czech | 90.88 |
| Italian | 96.50 | Serbian | 90.62 |
| Arabic | 96.25 | Icelandic | 86.75 |
| Greek | 95.62 | | |

Table 2: Gender-only differences between masculine and feminine translations across languages, as assessed by an LLM-as-a-judge approach.

The prompts used for data generation and evaluation are provided in Appendix A.

## 4 Evaluation Setup & Analysis Approach

For the evaluation, we prepared a dataset in which each entry consisted of a source text ($s$) in one of three source languages (English, Turkish, Finnish) and, for each target language, two corresponding hypotheses: one where the occupation appeared in the translated masculine form ($t_{male}$) and another where it appeared in the translated feminine form ($t_{female}$). The two hypotheses were constructed to differ solely in the grammatical gender marking of the occupation, with all other aspects of the sentence kept identical. This dataset was then provided by the shared task organizers to the participating teams, who ran their metrics and returned their scores for each hypothesis in every source–target language pair, while the organizers themselves used a set of baseline metrics on our dataset too. These returned results form the basis of the subsequent analysis. The source languages were 3 (English, Turkish, Finnish) and the target languages were 11 (Arabic, Czech, Greek, Spanish, French, Icelandic, Italian, Portuguese, Russian, Serbian, Ukrainian), leading to a total of 33 distinct source–target language pairs.

The organizers returned results for 11 evaluation metrics, 3 baseline metrics (sentinel-cand (Perrella et al., 2024), sentinel-src (Perrella et al., 2024), and COMETKiwi22 (Rei et al., 2022)), and 8 submissions (UvA-MT (Wu and Monz, 2025), ranked-COMET (Maharjan and Shrestha, 2025), MetricX-25-QE (Juraska et al., 2025), MetricX-25 (Juraska et al., 2025), baseCOMET (Maharjan and Shrestha, 2025), Polycand-1 (Züfle et al., 2025), Polycand-2 (Züfle et al., 2025), and Polyic-3 (Züfle et al., 2025)). All metrics, except for UvA-MT, were run on the complete set of 33 source-target language pairs. The UvA-MT metric was evaluated only with English as the source language, covering all corresponding target languages. sentinel-src relies exclusively on the input, which results in identical scores being assigned to both gendered forms; therefore, it is excluded from our subsequent analyses

## 5 Results and Discussion

### 5.1 Experimental Setup

Let $M(s,t) \in [\min(M), \max(M)]$ be a translation quality metric, where $s$ denotes the source text, $t$ the translated text, and $M(s,t)$ returns a real-valued score within the metric's range. Our goal is to quantify the variation in $M$ when the translations differ *only* in the grammatical gender of an occupation, while all other elements of the text remain identical.

For each source sentence $s$, we consider two translations: $t_{\text{male}}$ (masculine form) and $t_{\text{female}}$ (feminine form). The *absolute difference* in metric scores is defined as:

$$\Delta_{\text{abs}} = \left| M(s, t_{\text{male}}) - M(s, t_{\text{female}}) \right|. \quad (1)$$

To facilitate comparability across different metrics, we define the *normalized difference* as:

$$\Delta_{\text{norm}}(\%) = \frac{\Delta_{\text{abs}}}{R} \times 100, \quad (2)$$

where $R = \max(M) - \min(M)$ is the *range* of the metric and $I = [\min(M), \max(M)]$ denotes the corresponding interval of observed values. Since in the general case the theoretical bounds of $M$ may be unknown, we instead rely on empirical estimates derived from our evaluation results, using as $\min(M)$ and $\max(M)$ the smallest and largest values observed for each metric on the challenge set. This approach ensures that all metrics are interpreted within a consistent, data-driven scale.

### 5.2 Results

Table 3 summarizes the performance of all evaluated metrics on our challenge set, reporting both absolute and normalized differences alongside their empirical ranges and intervals. A paired $t$-test[8] was conducted for each metric and language, showing that all differences between $M(s, t_{\text{male}})$ and $M(s, t_{\text{female}})$ were statistically significant in every case ($p < 0.05$).

---

[8]https://docs.scipy.org/doc/scipy/reference/generated/scipy.stats.ttest_rel.html

| Metric | $\Delta_{\text{norm}}(\%)$ | $\Delta_{\text{abs}}$ | $R$ | $I$ |
|---|---|---|---|---|
| UvA-MT[5] | 104.29 | 0.1314 | 0.126 | [-0.521, -0.395] |
| rankedCOMET | 71.09 | 0.0455 | 0.064 | [0.468, 0.532] |
| MetricX-25-QE | 11.93 | 0.5711 | 4.787 | [-8.253, -3.467] |
| *sentinel-cand | 9.70 | 0.0559 | 0.576 | [-0.237, 0.339] |
| MetricX-25 | 7.45 | 0.6151 | 8.260 | [-12.494, -4.234] |
| baseCOMET | 4.39 | 0.0058 | 0.132 | [0.435, 0.567] |
| *COMETKiwi22 | 4.13 | 0.0100 | 0.242 | [0.623, 0.865] |
| Polycand-1 | 3.72 | 0.6023 | 16.184 | [76.655, 92.839] |
| Polycand-2 | 3.36 | 0.8590 | 25.549 | [68.230, 93.779] |
| Polyic-3 | 3.10 | 0.6197 | 19.985 | [74.547, 94.533] |

Table 3: Absolute and normalized differences for each metric, along with their empirical range $R$ and interval $I$ on the challenge set. $\Delta_{\text{abs}}$ and $\Delta_{\text{norm}}$ are reported as averages over all source sentences and language pairs. Baseline metrics are indicated with an asterisk (*).

All evaluated metrics show a measurable difference between masculine and feminine translations. Notably, the magnitude of these differences, both in absolute and normalized terms, varies considerably across metrics. For example, UvA-MT and ranked-COMET display the largest normalized differences (above 70%), suggesting higher sensitivity to gender variation, whereas metrics such as Polyic-3 and Polycand-2 register much smaller relative changes. While, the empirical ranges $R$ and intervals $I$ differ widely among metrics, reflecting substantial variation in score scales and distributional properties, normalization allows us to fairly compare the different metrics.

**Analysis per Language.** Table 4 presents the average normalized differences (± standard deviation) across all evaluation metrics for each source–target pair. Across most targets, English as the source leads to higher average differences than Finnish or Turkish. In certain cases, such as Arabic, Czech, Icelandic, Italian, Russian, Serbian, and Ukrainian, the source language changes the difference score substantially (>8%). In contrast, for Greek, Spanish, French, and Portuguese, the results are more stable regardless of the source. This suggests a systematic source-language effect. One possible explanation is that English, as a natural-gendered language, exhibits more extensive gender marking than Finnish or Turkish, which are largely genderless languages, thereby increasing the likelihood that occupational gender associations can be stronger and leak into the evaluation metrics. However, this does not explain why this is not true for all target languages, so further investigation is needed to disambiguate these findings.

When focusing on the target languages, the ranking is stable: Arabic, Russian, and Icelandic yield the highest values; Czech, Italian, Ukrainian, and Serbian follow; French, Greek, Spanish, and Portuguese are consistently lower. This stability across different sources points to target-language characteristics as a dominant factor. Standard deviations parallel the ranking of the averages—larger where averages are higher (especially with English as the source) and smaller where they are lower—indicating greater item-level variability when differences are larger. In high-difference pairs, this variability likely reflects greater diversity in how specific items respond to the different evaluation metrics. Conversely, in low-difference pairs, more uniform behavior is observed across items. Overall, the analysis highlights two main points: while the inherent characteristics of the target language play a dominant role in determining the magnitude of differences, selecting English as the source systematically amplifies these effects, leading to both higher average values and greater variability across samples.

| Source \ Target | English | Finnish | Turkish |
|---|---|---|---|
| Arabic | $25.39_{\pm 40.28}$ | $16.82_{\pm 32.45}$ | $18.09_{\pm 35.52}$ |
| Czech | $22.57_{\pm 34.42}$ | $13.38_{\pm 21.14}$ | $13.76_{\pm 23.01}$ |
| Greek | $11.55_{\pm 18.48}$ | $10.89_{\pm 18.55}$ | $10.97_{\pm 19.07}$ |
| Spanish | $11.79_{\pm 19.68}$ | $12.32_{\pm 22.05}$ | $12.32_{\pm 22.99}$ |
| French | $9.47_{\pm 14.47}$ | $8.95_{\pm 13.77}$ | $9.13_{\pm 14.60}$ |
| Icelandic | $26.10_{\pm 32.77}$ | $16.62_{\pm 19.58}$ | $16.55_{\pm 20.51}$ |
| Italian | $22.14_{\pm 35.30}$ | $12.59_{\pm 21.41}$ | $12.27_{\pm 21.42}$ |
| Portuguese | $11.62_{\pm 20.12}$ | $11.41_{\pm 20.14}$ | $11.58_{\pm 21.51}$ |
| Russian | $25.63_{\pm 38.98}$ | $16.38_{\pm 29.50}$ | $15.49_{\pm 27.28}$ |
| Serbian | $20.17_{\pm 33.76}$ | $11.64_{\pm 20.18}$ | $11.68_{\pm 21.05}$ |
| Ukrainian | $21.64_{\pm 34.86}$ | $12.93_{\pm 22.08}$ | $12.90_{\pm 21.90}$ |

Table 4: Average $\Delta_{\text{norm}}(\%)$ with standard deviation for each source-target pair across all models.

---

[5]Results only with English as the source language.

| Occupation | sentinel-cand | COMETKiwi22 | MetricX-25 | UvA-MT | rankedCOMET | Polycand-2 | Polyic-3 | MetricX-25-QE | baseCOMET | Polycand-1 | AVERAGE |
|---|---|---|---|---|---|---|---|---|---|---|---|
| | | | | | $\overline{M(s, t_{male})} > \overline{M(s, t_{female})}$ | | | | | | |
| 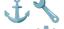 | +9.7% | +11.9% | +13.1% | +58.0% | +6.1% | +13.6% | +15.5% | +27.0% | +5.9% | +18.8% | +18.0% |
| 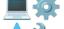 | +15.0% | +8.8% | +12.5% | +34.4% | +14.3% | +16.2% | +18.1% | +25.4% | +11.9% | +16.9% | +17.3% |
| 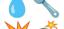 | +9.4% | +9.3% | +22.5% | +39.6% | +6.4% | +11.1% | +11.7% | +33.9% | +5.3% | +15.1% | +16.4% |
| 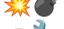 | +11.5% | +10.3% | +19.9% | +26.7% | +5.6% | +13.5% | +15.4% | +33.3% | +6.7% | +17.5% | +16.0% |
| 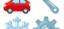 | +11.3% | +10.1% | +15.6% | +39.8% | +7.3% | +13.1% | +13.6% | +25.1% | +6.9% | +17.0% | +16.0% |
| 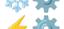 | +9.3% | +8.1% | +16.7% | +36.4% | +7.5% | +14.6% | +13.8% | +29.5% | +7.2% | +16.5% | +16.0% |
| 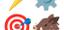 | +13.5% | +11.5% | +13.2% | +31.3% | +9.8% | +12.7% | +14.2% | +24.7% | +10.1% | +15.4% | +15.7% |
| 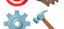 | +6.4% | +8.5% | +15.8% | +41.8% | +5.1% | +13.0% | +11.4% | +36.0% | +4.8% | +12.9% | +15.6% |
| 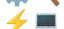 | +9.5% | +10.2% | +17.9% | +34.0% | +4.4% | +13.5% | +15.2% | +28.6% | +5.2% | +16.7% | +15.5% |
| 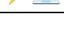 | +15.9% | +9.2% | +11.4% | +24.6% | +10.6% | +15.3% | +21.1% | +20.6% | +9.2% | +15.9% | +15.4% |
| | | | | | $\overline{M(s, t_{male})} < \overline{M(s, t_{female})}$ | | | | | | |
| 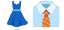 | -2.5% | -6.1% | -13.8% | -22.4% | +1.8% | +0.7% | -3.1% | -25.2% | +0.8% | -2.7% | -7.2% |
| 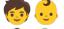 | -3.5% | -3.9% | -13.7% | -24.3% | -0.8% | -1.0% | +0.4% | -21.7% | -0.7% | +0.6% | -6.9% |
| 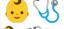 | -1.0% | -5.8% | -18.2% | -5.3% | +4.0% | +1.7% | +0.5% | -28.1% | +3.3% | +1.7% | -4.7% |
| 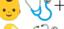 | -2.1% | -2.6% | -21.3% | -4.0% | +6.0% | +1.6% | -0.6% | -28.1% | +5.0% | +4.2% | -4.2% |
| 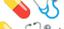 | -1.4% | -2.3% | -10.6% | -8.7% | -0.2% | +1.8% | +2.6% | -15.2% | -0.3% | +1.3% | -3.3% |
| 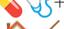 | -1.2% | -3.8% | -7.5% | -14.8% | +0.6% | +1.9% | +2.3% | -10.5% | +0.1% | +2.1% | -3.1% |
| 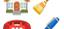 | -0.1% | +1.8% | -11.4% | -8.7% | +1.9% | +0.9% | +0.1% | -14.7% | +1.2% | +1.5% | -2.7% |
| 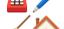 | -1.1% | +0.8% | -5.7% | -2.3% | +1.1% | +1.3% | +0.8% | -11.4% | +0.9% | +1.8% | -1.4% |
| 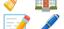 | +0.1% | +1.2% | -10.3% | -1.9% | +3.9% | +2.6% | +2.4% | -14.1% | +3.4% | +2.1% | -1.1% |
| 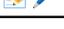 | -0.6% | +1.1% | -7.5% | -5.9% | +3.8% | +3.2% | +4.7% | -13.1% | +3.3% | +3.3% | -0.8% |

Table 5: $\Delta_{\text{norm}}(\%)$ per occupation across all evaluated metrics. The occupations referenced in this table are listed in Table 7 along with the respective ISCO codes.

**Analysis per Occupation.** For this analysis, it is necessary to define not only the *strength* of the effect, captured by the absolute magnitude $\Delta_{\text{norm}}(\%)$, but also its *direction*. Specifically, a "+" is assigned when $\overline{M(s, t_{\text{male}})} > \overline{M(s, t_{\text{female}})}$ (where the overline denotes the mean value), indicating that the metric tends to favor the masculine form, whereas a "–" is assigned when $\overline{M(s, t_{\text{male}})} < \overline{M(s, t_{\text{female}})}$, indicating a tendency to favor the feminine form.

Table 5 presents the top ten occupations with the largest positive and negative normalized differences $\Delta_{\text{norm}}(\%)$ across all evaluated metrics, ranked by their average value across all languages. Occupations are identified using their corresponding ISCO-08 codes, providing a standardized reference for each job category.

The occupations appearing in the positive segment of the table (e.g., ISCO 3151 "Ships' Engineers", ISCO 7126 "Plumbers and Pipe Fitters", ISCO 7542 "Shotfirers and Blasters") are predominantly roles that are culturally and historically associated with men and are often perceived as physically demanding or male-dominated. In contrast, the negative segment features occupations such as ISCO 5241 "Fashion and Other Models", ISCO 5311 "Child Care Workers", and ISCO 3222 "Midwifery Professionals", which are stereotypically viewed as female-oriented professions. This pattern suggests that the evaluated metrics may reproduce gender stereotypes in their outputs, ultimately reinforcing such biases in MT systems.

It is also noteworthy that the magnitude of the positive differences is, in absolute terms, generally greater than that of the negative differences. For example, the tenth-highest positive difference toward the masculine form substantially exceeds the highest negative difference toward the feminine form. This asymmetry indicates an overall tendency of the evaluated metrics to favor masculine forms in translation.

## 6 Gender Density Analysis

Different human-evaluation approaches of translation penalize errors (such as using the wrong gender) differently: direct assessment applies a conceptual approach, penalizing each error only once (Graham et al., 2017), whereas MQM (Lommel et al., 2013) penalizes every occurrence of the error in a text. To better understand how QE metrics capture gender related aspects, we analyse the correlation between the bias metric (normalized difference score $\Delta_{\text{norm}}(\%)$) and gender density in the text. We define *gender density* as the normalized count of gendered words in the text. Since the two texts ($t_{male}, t_{female}$) are identical except for the gendered words, we calculate the gender density as the proportion of differing words to the average text length (in words):

$$\text{gender density} = \frac{\text{worddiff}(t_{male}, t_{female})}{\text{average}(\text{len}(t_{male}), \text{len}(t_{female}))}$$

| Metric | English | Finnish | Turkish |
|---|---|---|---|
| *COMETKiwi22 | 0.17 | 0.21 | 0.22 |
| MetricX-25 | 0.02 | 0.09 | 0.08 |
| MetricX-25-QE | 0.06 | 0.17 | 0.19 |
| Polycand-1 | 0.18 | 0.22 | 0.22 |
| Polycand-2 | 0.17 | 0.21 | 0.20 |
| Polyic-3 | 0.12 | 0.16 | 0.18 |
| UvA-MT | 0.47 | — | — |
| baseCOMET | 0.20 | 0.21 | 0.22 |
| rankedCOMET | 0.09 | 0.10 | 0.11 |
| *sentinel-cand | 0.02 | 0.02 | 0.02 |

Table 6: Pearson correlation coefficients between gender density and metrics' bias. All metrics show a positive correlation and all correlations are statistically significant ($p < 0.05$). Baseline metrics are indicated with an asterisk (*).

As presented in Table 6, we observe a positive correlation between metric bias and gender density for all metrics with varying correlation strength, i.e. *texts with a higher number of gendered words result in more biased metric evaluations*. Among the evaluated metrics, Polycand metrics show the strongest correlations overall to gender density. COMET-based metrics (COMETKiwi22, baseCOMET, rankedCOMET) also show strong correlations, while MetricX-25 and MetricX-25-QE consistently show weak correlations. Sentinel-cand metric shows minimal correlations, suggesting it may be less sensitive to gender density. Detailed correlation plots for each metric are in appendix C. The findings suggest that the metrics apply a cumulative preference and penalty for gender in a text.

## 7 Conclusions

We presented GAMBIT+, a large-scale, fully parallel challenge set containing paired masculine and feminine translations of gender-ambiguous occupations across 33 source–target language pairs, and used it to benchmark 10 QE metrics. Our evaluation revealed consistent, statistically significant score shifts driven solely by grammatical gender—often more pronounced in certain language pairs and occupations—with a clear overall tendency to favor masculine forms. These findings indicate that current QE metrics are not gender-fair and should be systematically audited and calibrated.

Future work will broaden the scope of our analysis to include a wider range of QE metrics, with a focus on identifying specific characteristics that make them more susceptible to gender bias. We also plan to investigate the interplay between MT systems and QE metrics, exploring how system outputs and metric evaluations align or diverge in terms of gender bias. We also plan to extend GAMBIT+ with translations of varying quality to investigate whether translation quality influences the biases of QE metrics—for example, whether poor translations make gender distinctions more or less apparent due to incorrect gender agreements across the sentence. Finally, we aim to extend the use of GAMBIT+ beyond MT, testing its applicability and value in evaluating gender bias in other natural language generation tasks.

## Acknowledgments

This work is part of the UTTER project, supported by the European Union's Horizon Europe research and innovation programme via grant agreement 101070631, and by the FCT project "OptiGov", ref. 2024.07385.IACDC (DOI 10.54499/2024.07385.IACDC), funded by the PRR under the measure RE-C05-i08.m04. Giuseppe Attanasio and Chrysoula Zerva are also supported by the Portuguese Recovery and Resilience Plan through projects C645008882-00000055 (Center for Responsible AI) and UID/50008: Instituto de Telecomunicações. Chrysoula Zerva is also supported by an unrestricted gift from Google (Google Research Scholar). Wafaa Mohammed acknowledges travel support from ELIAS (GA no 101120237).

## A  Prompts

Below is the prompt designed for the translation task. In this context, the variable {gender} can take the value "male" or "female," {occupation} refers to the specific occupation under study as defined by the GAMBIT framework, and {lang} denotes the target language. The prompt instructs the translation model to produce a faithful and semantically equivalent rendering of the source text, adapting all gendered references to match the specified {gender} form for the given {occupation}.

---

**Gendered Translation Prompt**

You are an expert professional translator. Translate the following English text into {lang}.

**Important instructions:**

- Any reference to the word "{occupation}" must be translated using the {gender} form in {lang}.

- Be extremely careful to preserve the exact meaning, tone, and semantics of the original text.

- Do **not** modify, omit, or add any information other than adapting the gender reference to {gender} for the {occupation}.

- Return **only the translation**, without any additional comments, notes, or explanations.

Text:

{text}

Translation:

---

Below is the prompt used for the gender-adaptation procedure in translations. In this setting, the variable {gender} assumes the opposite value from that assigned in the initial stage of the workflow-that is, if the initial assignment is "male," it is here set to "female," and vice versa. The variable {occupation} corresponds to the occupation label provided in the GAMBIT dataset, while {lang} specifies the language into which the translation is to be rendered. The instructions require the translation model to maintain complete semantic fidelity to the source text, altering only the gendered form of the {occupation} in accordance with the designated {gender}, without introducing, removing, or rephrasing any other content.

---

**Gender Adaptation Prompt**

You are an expert professional translator. Translate the following English text into {lang}.

***Important instructions:***

- Any reference to the word "occupation" must be translated using the {gender} form in {lang}.

- Be extremely careful to preserve the exact meaning, tone, and semantics of the original text.

- Do ***not*** modify, omit, or add any information other than adapting the gender reference to {gender} for the "{occupation}".

- Return ***only the translation***, without any additional comments, notes, or explanations.

Text:

{text}

Translation:

---

For languages without grammatical gender, namely Turkish and Finnish, no gender adaptation process is applied. In these cases, only the standard translation prompt is employed, as provided below, ensuring that occupational and role references remain neutral in accordance with the structural characteristics of the target language.

---

**Genderless Translation Prompt**

You are an expert professional translator. Translate the following English text into lang.

Important instructions: - Pay close attention to the gender of any occupations or role titles. - If the gender is not clear or not mentioned, do not assume or infer it; keep the translation gender-neutral or ambiguous as appropriate in the target language. - Return only the translation, without any additional comments, notes, or explanations.

Text:

{text}

Translation:

---

For the evaluation phase, the following prompt was employed to compare two texts in the target language. Its primary function was to determine whether any differences between the two texts were solely attributable to the explicit mention of the gender of the specified occupation or whether other differences were present. Text 1 and Text 2 correspond to the two translations of the same source text, each rendered with a different specified gender for the occupation under study.

> **Evaluation Prompt**
>
> You are a linguistic comparison assistant. Analyze the following two {lang} texts and identify the differences between them.
>
> If the only differences are related to the explicit mention of the gender of the occupation occupation_title (e.g., masculine vs. feminine forms of the occupation {occupation_title}), respond with "yes".
>
> If you find any other type of difference, respond with "no" and list all the differences you found.
>
> Return only the answer, without any additional comments, notes, or explanations, especially when the answer is "yes".
>
> Text 1:
>
> {text1}
>
> Text 2:
>
> {text2}

## B  ISCO Code–Occupation Mapping

Table 7 provides the mapping between the ISCO codes referenced in Table 1 and their corresponding occupational titles as defined in the ISCO-08 classification.

| ISCO Code | ISCO Name | Emojis |
|---|---|---|
| 3151 | Ships' Engineers | ⚓🔧 |
| 3513 | Computer Network and Systems Technicians | 💻⚙ |
| 7126 | Plumbers and Pipe Fitters | 💧🔧 |
| 7542 | Shotfirers and Blasters | 💥💣 |
| 7231 | Motor Vehicle Mechanics and Repairers | 🚗🔧 |
| 7127 | Air Conditioning and Refrigeration Mechanics | ❄⚙ |
| 7421 | Electronics Mechanics and Servicers | ⚡⚙ |
| 6224 | Hunters and Trappers | 🎯🦡 |
| 7213 | Sheet Metal Workers | ⚙🔨 |
| 3114 | Electronics Engineering Technicians | ⚡💻 |
| 5241 | Fashion and Other Models | 👗👔 |
| 5311 | Child Care Workers | 👶👼 |
| 2222 | Midwifery Professionals | 👶🩺 |
| 3222 | Midwifery Associate Professionals | 👶🩺+ |
| 2221 | Nursing Professionals | 💊🩺 |
| 3221 | Nursing Associate Professionals | 💊🩺+ |
| 5151 | Cleaning and Housekeeping Supervisors | 🏠🧹 |
| 4226 | Receptionists (general) | ☎🖊 |
| 9111 | Domestic Cleaners and Helpers | 🧹🏠 |
| 4120 | Secretaries (general) | 📝🖊 |

Table 7: Mapping of emojis from Table 5 to their corresponding occupational titles and the respective ISCO codes.

## C  Gender-Density Correlation Plots

Figure 1 shows the correlation plots of gender density to normalized score difference for each metric. Plots are shown for English as a source language. Other source languages show the same trend.

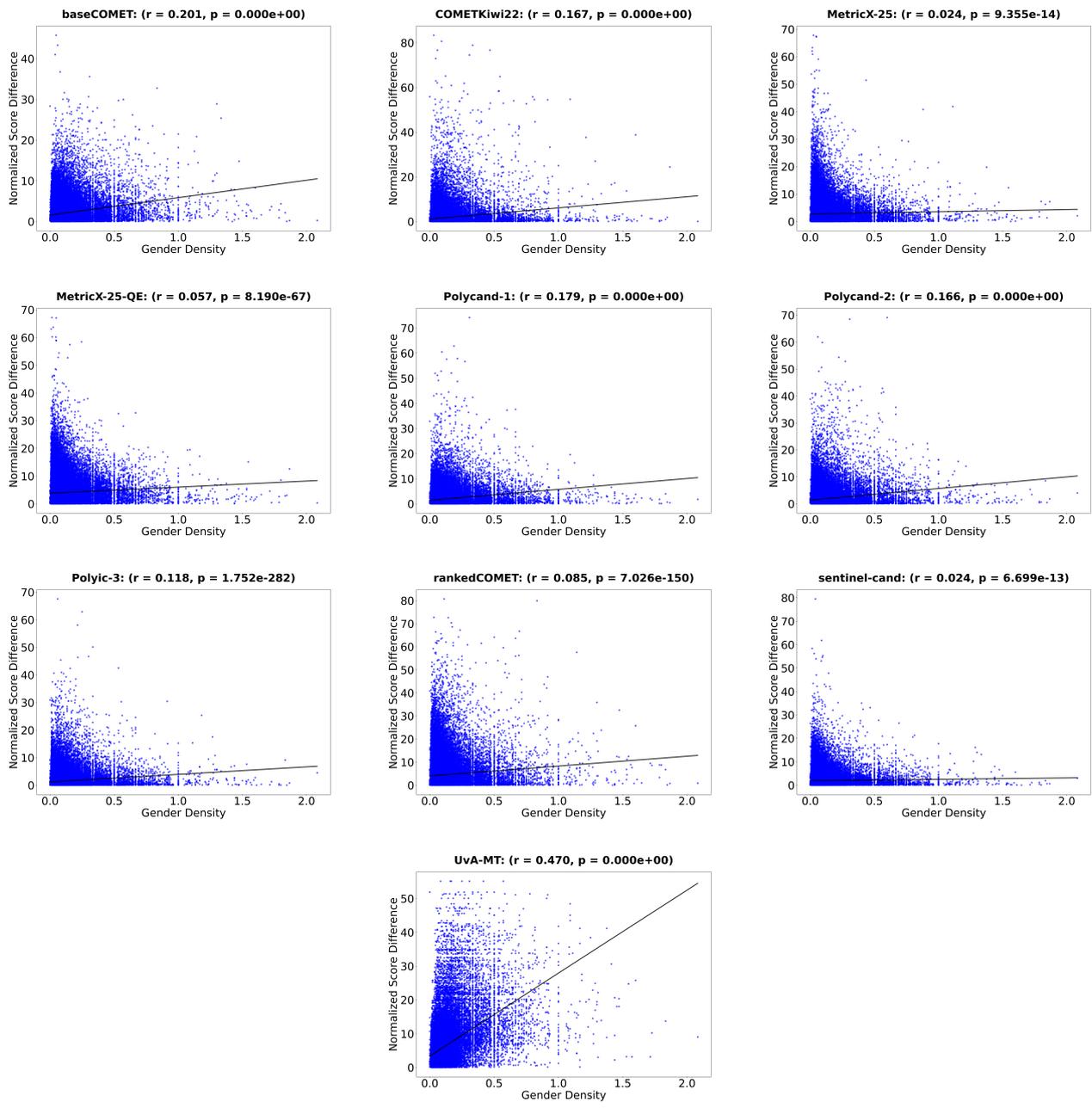

Figure 1: Gender density to normalized score difference correlation for all metrics.